\documentclass[runningheads]{llncs}
\usepackage{graphicx}
\graphicspath{{figs/}}
\usepackage{xcolor}

\usepackage{multirow}
\usepackage{tabularx}
\usepackage{booktabs} 

\usepackage[misc,geometry]{ifsym}

\begin{document}
\title{ICDAR 2021 Competition on Components Segmentation Task of Document Photos\thanks{Supported by University of Pernambuco, CAPES and CNPq.}}
\titlerunning{ICDAR 2021 Competition on CSTDP}
%

\author{Celso A. M. Lopes Junior\inst{1}\orcidID{0000-0003-1356-5759} \and
Ricardo B. das Neves Junior\inst{1}\orcidID{0000-0001-9538-6505} \and
Byron L. D. Bezerra\inst{1}(\Letter)\orcidID{0000-0002-8327-9734} \and
Alejandro H. Toselli\inst{2}\orcidID{0000-0001-6955-9249} \and
Donato Impedovo\inst{3}\orcidID{0000-0002-9285-2555}}
\authorrunning{C. A. M. Lopes Junior et al.}
%
\institute{Polytechnic School of Pernambuco - University of Pernambuco, Pernambuco, Brazil \email{\{camlj, rbnj\}@ecomp.poli.br}, \email{byron.leite@upe.br} \and
PRHLT Research Center, Universitat Politécnica de Valéncia, Camí de Vera, s/n, 46022 Valéncia, Spain
\email{a.toselli@northeastern.edu} \and 
Department of Computer Science, University of Bari, Bari, Italy \\
\email{donato.impedovo@uniba.ite}}
\maketitle              
\begin{abstract}

  This paper describes the short-term competition on ``Components
  Segmentation Task of Document Photos'' that was prepared in the
  context of the ``16th International Conference on Document Analysis
  and Recognition'' (ICDAR 2021).
  This competition aims to bring together researchers working on the
  filed of identification document image processing and provides them
  a suitable benchmark to compare their techniques on the component
  segmentation task of document images.
  Three challenge tasks were proposed entailing different segmentation
  assignments to be performed on a provided dataset. The collected
  data are from several types of Brazilian ID documents, whose
  personal information was conveniently replaced.
  There were 16 participants whose results obtained for some or all
  the three tasks show different rates for the adopted metrics, like
  ``Dice Similarity Coefficient'' ranging from $0.06$ to
  $0.99$. Different Deep Learning models were applied by the entrants
  with diverse strategies to achieve the best results in each of the
  tasks.
  Obtained results show that the current applied methods for solving
  one of the proposed tasks (document boundary detection) are already
  well stablished. However, for the other two challenge tasks (text
  zone and handwritten sign detection) research and development of
  more robust approaches are still required to achieve acceptable
  results.
  
\end{abstract}

\section{Introduction}
\label{section:introduction}

This paper describes the short-term competition on ``Components
Segmentation Task of Document Photos'' organized in the context of the
``16th International Conference on Document Analysis and Recognition''
(ICDAR 2021).

The traffic of identification document images (ID document) through
digital media is already a common practice in several countries. A
large amount of data can be extracted from these images through
computer vision and image processing techniques, as well machine
learning approaches.
The extracted data can serve for different purposes, such as names and
dates retrieved from text fields to be processed by OCR systems or
extracted handwritten signatures to be checked by biometric systems,
in addition to other characteristics and patterns present in images of
identification documents.
%
Actually there are few developed image processing applications that
focus on the treatment of ID document images, specially related with
image segmentation of ID documents.

The main goal of this contest was to stimulate researchers and
scientists in the search for new techniques of image segmentation for
the treatment of these ID document images.
The availability of an adequate experimental dataset of ID documents
is another factor of great importance, given that there are few of
them free-available to the community~\cite{de2020bid}. Due to privacy
constraints, for the provide dataset all text with personal
information was synthesized with fake data. Furthermore, the original
signatures were substituted by new ones collected randomly from
different sources.
All these changes were performed by well-designed algorithms and
post-processed by humans whenever needed to keep the real-world
conditions as much as possible.

\section{Competition Challenges and Dataset Description}
\label{section:challenge_and_dataset}

In this contest, entrants have to employ their segmentation techniques
on given ID document images according to tasks defined for different
levels of segmentation.
Specifically, the goal here is to evaluate the quality of the applied
image segmentation algorithms to ID document images acquired by mobile
cameras, where several issues may affect the segmentation task
performance, such as: location, texture and background of the
document, camera distance to the document, perspective distortions,
exposure and focus issues, reflections or shadows due to ambient light
conditions, among others.

To evaluate tasks entailing different segmentation levels, the
following three challenges have to be addressed by the participants as
described below.

\subsection{Challenge Tasks}

\subsubsection*{1st Challenge - Document Boundary Segmentation:}
\begin{figure}[!htb]
  \centering
  \includegraphics[width=.48\textwidth]{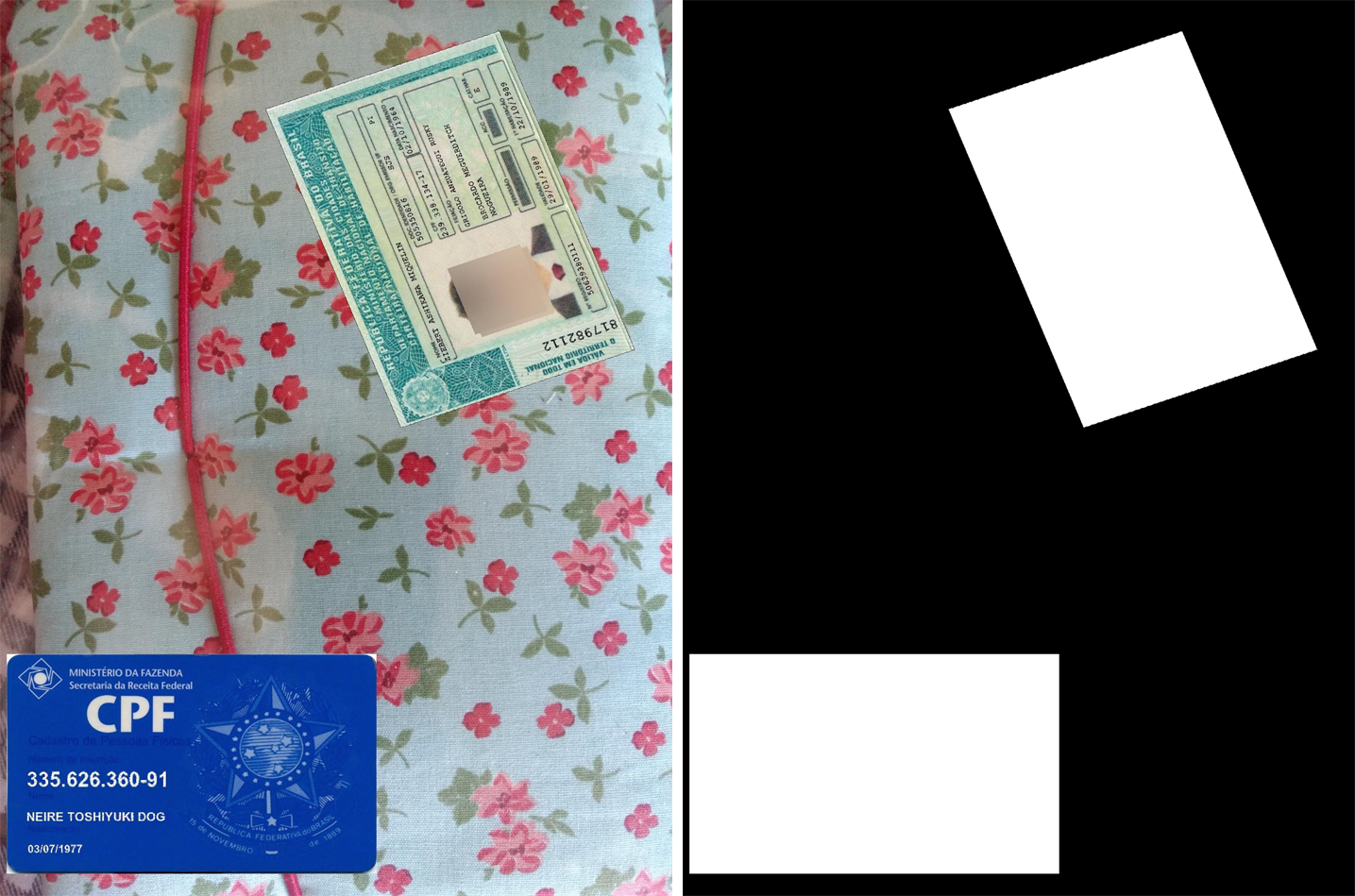}%
  \hfill%
  \raisebox{-.4ex}{\includegraphics[width=.48\textwidth]{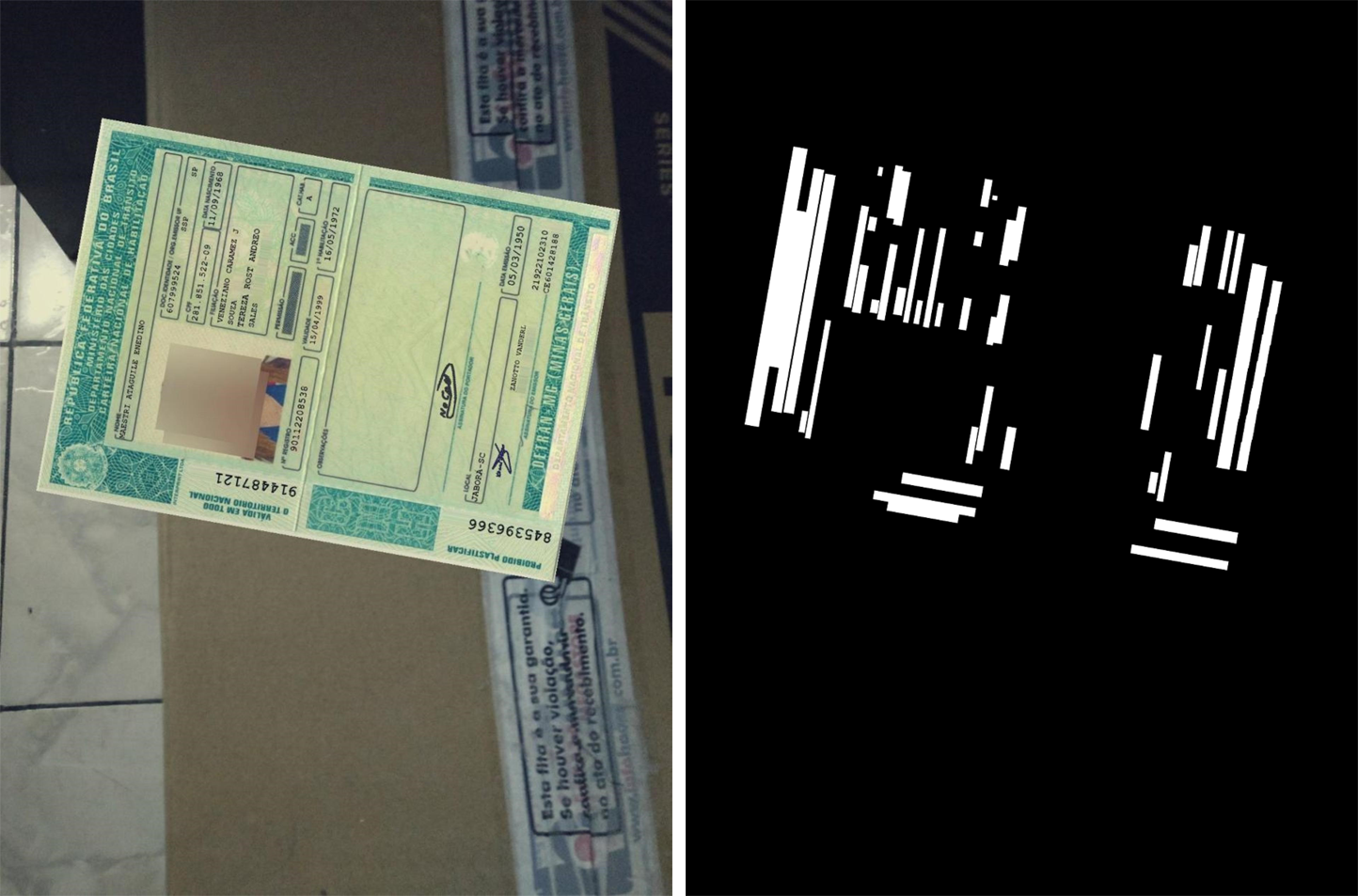}}
  \caption{Left: example of images containing a document and their
    boundary detection output images. Right: an example of a document
    image and its output image with detected text regions masked.}
  \label{fig1}
\end{figure}

The objective of this challenge is to develop boundary detection
algorithms for different kinds of documents~\cite{das2020fast}. The
entrants should develop an algorithm that takes as input an image
containing a document, and return a new image of the same size with
the background in black pixels and the region occupied by the document
in white pixels. Figure~\ref{fig1}-left shows an example of this
detection process.

\subsubsection*{2nd Challenge - Zone Text Segmentation:}

This challenge encourages the development of algorithms for automatic
text detection in ID documents~\cite{das2020fast}. The entrants have
to develop an algorithm capable of detecting text patterns in the
provided set of images; that is, to process an image of a document
(without background), and return a new image of the same size with
non-interest regions in black pixels and regions of interest (text
regions) in white pixels. This detection process is illustrated in
Figure~\ref{fig1}-right.



\subsubsection*{3rd Challenge - Signature Segmentation:}

\begin{figure}[!htb]
  \centering
  \includegraphics[width=.7\textwidth]{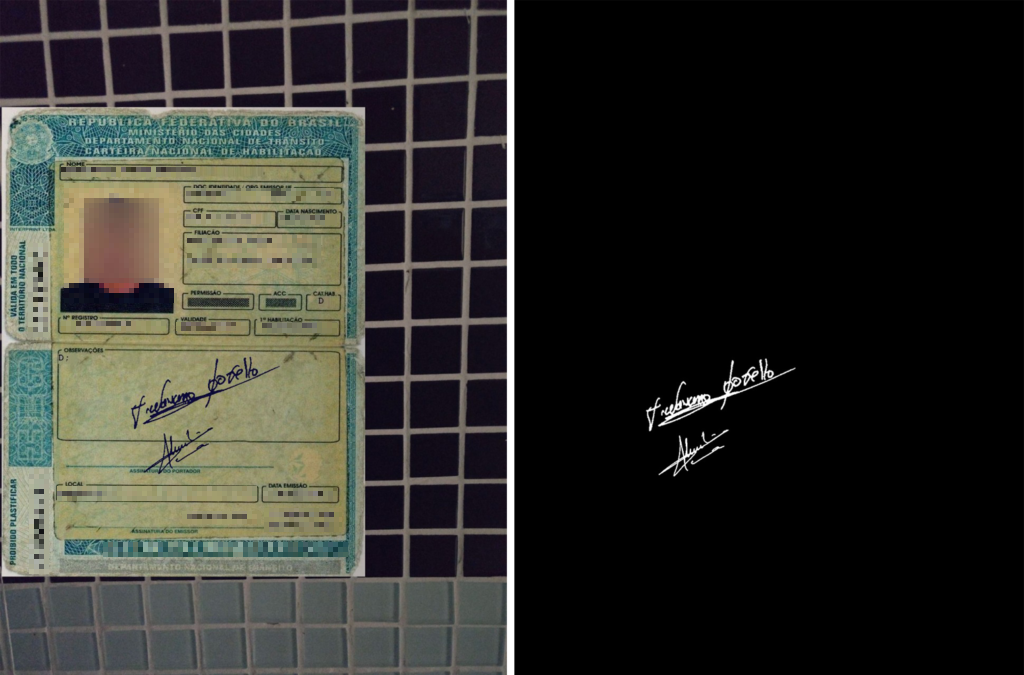}
  \caption{Example of a document image and its output image with
    a detected handwritten signature masked.}
  \label{fig3}
\end{figure}

This challenge aims at developing algorithms to detect and segment
handwritten signatures on ID
documents~\cite{junior2020fcn,silva2019speeding}. Given an image of a
document, the model or technique applied should return an image with
the same size of the input image with the handwritten signature
strokes as foreground in white pixels, and everything else in black
pixels. Figure~\ref{fig3} displays an example of this detection.

\subsection{ID Document Image Dataset}

For this competition, the provided dataset is composed of thousands of
images of ID documents from the following Brazilian document types:
National Driver's License (CNH), Natural Persons Register (CPF) and
General Registration (RG).
The document images were captured through different cell phone cameras
at different resolutions. Documents that appear in the images are
characterized by having different textures, colors and lighting, on
different real-world backgrounds with non-uniform patterns.

The images are all in RGB color mode and PNG format. The ground truth
of this dataset
is different for each challenge. For the first challenge, the interest
regions are the document areas themselves, while the printed text areas
are the ones for the second challenge. For the third challenge, we are
only interested in the pixels of detected handwritten signature
strokes. As commented before, for all the challenges, the interest
regions/pixels are represented by white pixel areas, while
non-interest ones in black pixel areas.

As personal information contained in the documents cannot be made public, the original data was replaced with synthetic data. We generate fake information for all fields present in the document (such as name, date of birth, affiliation, number of document, among others), as described in~\cite{de2020bid}.

On the other hand, handwritten signatures were acquired from the
MCYT~\cite{ortega2003mcyt} and GPDS~\cite{ferrer2014static} datasets, which were synthetically
incorporated into the images of ID documents. 

Both the ID document images and the corresponding ground truth of the
dataset were manually verified, and they are intended to be used for
model training and for computing evaluation
metrics. Table~\ref{tab:dataset} summarizes the basic statistics of
the dataset. The whole dataset is available for the research community under request (visit: \url{https://icdar2021.poli.br/}).


\begin{table}[!htb]
  \centering
  \extrarowheight=0.1pt
  \tabcolsep=5pt
  \caption{Statistics of the dataset and experimental partition used
    in the competition, where C1, C2, and C3 corresponds to 1st, 2nd and 3rd challenge, respectively.}
  \label{tab:dataset}
  \begin{tabular}{l|rr|rr|rr}
    \toprule
    Number of: & Train-C1 & Test-C1 & Train-C2 & Test-C2 & Train-C3 & Test-C3 \\
    \midrule
    Images & 15,000 & 5,000 & 15,000 & 5,000 & 15,000 & 5,000 \\
    \bottomrule
  \end{tabular}
\end{table}


\section{Competition Protocol}
\label{section:competition_protocol}


As introduced in the previous section, this competition presents three challenges:
\begin{itemize}
    \item 1st Challenge: Document Boundary Segmentation
    \item 2nd Challenge: Zone Text Segmentation
    \item 3rd Challenge: Signature Segmentation
\end{itemize}

Competitors can decide how many tasks they want to compete for, choosing between one or more challenges. In addition to the competition data, it was allowed to use additional datasets for model training. In this sense, some participants used the BID Dataset~\cite{de2020bid}, and Imagenet~\cite{deng2009imagenet} for improving the prediction of their approaches. 

Participants could also use solutions from a challenge task as support to develop solutions for other challenges. Some participants used image segmentation predictions of the 1st challenge to remove background from images
before they were used for the 2nd and 3rd challenge tasks.

Participants could also use pre or post-image processing techniques to develop their solutions.

\subsection{Assessment Methods}


The Structural Similarity Index (SSIM) metric was employed in the evaluation process. However, the results obtained using SSIM were very similar in all systems and techniques. Therefore we did not report their results in this work.

The following similarity metrics were used for evaluating the
approaches applied by the participants:

\subsubsection{Dice Similarity Coefficient (DSC)}

The Dice Similarity Coefficient (DSC), presented by Equation \ref{dsc_equation} is a statistical metric proposed by Dice, Lee R \cite{dice1945measures}, which can be used to measure the similarity between two images.

\begin{equation}
DSC = \frac{2 |X \cap Y|}{|X| + |Y|}
\label{dsc_equation}
\end{equation} where $|X|$ and $|Y|$ are elements of the ground truth and segmented image, respectively.

\subsubsection{Scale Invariant Feature Transform (SIFT)}

The scale-invariant feature transform (SIFT) is a feature extraction algorithm, proposed by David Lowe \cite{lowe1999object}, that describe local features in images.

The feature extraction performed by SIFT consists in the detection of Key-Points at the image. These key-points denote information related to the regions of interest (represented by the white pixels) of the ground truth and the segmented image. Once mapped, the key-points of the ground truth images are stored and compared with the key-points identified in the predicted image. From this comparison, it is possible to define a similarity rate.



%

\section{Briefly Description of the Entrant's Approaches}
\label{section:approaches}

In total there were 16 research groups from different countries that
have participated in the contest. The entrants submitted their
segmentation hypotheses computed on the corresponding test sets for
some or all of the three challenge tasks.
The main characteristics of the approaches used by each entrant in
each challenge task are described below.

\begin{enumerate}\itemsep=.5em
\item \textbf{58 CV Group (58 Group, China) -} \textbf{Task 1:}
  The approach is based on ``Efficient and Accurate Arbitrary-Shaped
  Text Detection with Pixel Aggregation Network''
  (PANNet)\,\cite{wang2019efficient} with improvements for edge
  segmentation. For model training, the segmentation effect has been
  improved by employing two different loss functions: the edge
  segmentation loss and the focal loss\,\cite{lin2017focal}.

\item \textbf{Ambilight Lab Group (NetEase Inc., China) -}
  \textbf{Task 1:} Document boundary detection has been carried out
  by using the composed model
  HRnet+OCR+DCNv2\,\cite{sun2019high,yuan2019object,zhu2019deformable}.
  \textbf{Task 2:} On the segmentation results in Task 1, a text
  angle classify model based on ResNet-50\,\cite{he2016deep} and a
  text segmentation model HRnet+OCR+DCNv2 were applied. \textbf{Task
    3:} On the segmentation results in Task 1, an approach based on
  YOLO-v5%
  \footnote{\url{https://zenodo.org/record/4679653\#.YKkfQSUpBFQ}} %
  for signature detection and HRnet+OCR+DCNv2 for signature
  segmentation were applied.

\item \textbf{Arcanite Group (Arcanite Solutions LLC, Switzerland) -}
  \textbf{Tasks 1, 2 \& 3:} U-Net\,\cite{ronneberger2015u} is employed
  as model architecture, where the encoder is based on
  ResNet-18\,\cite{he2016deep} and ImageNet\,\cite{deng2009imagenet},
  while the decoder is composed basically by transpose convolutions
  for up-sampling the
  outputs. Cross-entropy\,\cite{zhang2018generalized} and
  Dice\,\cite{sudre2017generalised} was used as loss functions in the
  training process.

\item \textbf{Cinnamon AI Group (Hochiminh City Univ. of Tech.,
    Vietnam) -} \textbf{Tasks 1, 2 \& 3:} Methods in these Tasks
  employ U-Net\,\cite{ronneberger2015u} as model architecture with
  Resnet-34\,\cite{he2016deep} backbone. This model has been trained
  to resolve the assignment proposed in each Task, where corresponding
  data augmentation has been applied for improving generalization and
  robustness.

\item \textbf{CUDOS Group (Anonymous) -} \textbf{Tasks 1, 2 \& 3:} The
  approaches are based on encoder-decoder
  SegNet\,\cite{badrinarayanan2017segnet} architecture and ``Feature
  Pyramid Network'' (FPN)\,\cite{lin2017feature} for object detection
  with MobileNetV2\,\cite{sandler2018mobilenetv2} backbone of the
  network.  Data augmentation and
  cross-entropy\,\cite{zhang2018generalized} loss were employed in the
  training of the network.

\item \textbf{Dao Xianghu light of TianQuan Group (CCB Financial
    Technology Co. Ltd, China) -} The approaches performed on the
  three tasks employ encoder-decoder model architectures. \textbf{Task
    1:} The approach uses two DL models (both based on ImageNet
  pretrained weights): DeepLabV3\,\cite{chen2017rethinking} (with
  ResNet-50 backbone) and Segfix\,\cite{yuan2020segfix} for
  refinement. \textbf{Task 2:} The applied approach is based on a
  model named DeepLabV3+\,\cite{chen2018encoder}. \textbf{Task 3:} In
  this case, the used approach is the
  Unet++\,\cite{zhou2018unet++}. In general, training of the models
  for the three tasks were carried out with data augmentation, using
  pretrained weights initialized on ImageNet, and SOTA (or
  FocalDice)\,\cite{lin2017focal} loss function.

\item \textbf{DeepVision Group (German Research Center for AI,
    Germany) -} \textbf{Tasks 1 \& 2}: The assignments in these Tasks
  were formulated as an instance segmentation problem. Basically, the
  approach uses Mask R-CNN\,\cite{he2017mask} for instance
  segmentation, with backbone of Resnet-50\,\cite{he2016deep}
  pretrained on ImageNet Dataset\,\cite{deng2009imagenet}.

\item \textbf{dotLAB Group (Universidade Federal de Pernambuco,
    Brazil) -} \textbf{Tasks 1, 2 \& 3}: The applied approaches are
  based on a network architecture called ``Self-Calibrated U-Net''
  (SC-U-Net)\,\cite{liu2020improving} which uses self-calibrated
  convolutions to learn better discriminative representations.

\item \textbf{NAVER Papago Group (NAVER Papago, South Korea) -}
  \textbf{Tasks 1, 2 \& 3}: The approaches are based on
  Unet++\,\cite{zhou2018unet++} with encoder based on
  ResNet\,\cite{he2016deep}. Training of models were conducted with
  data augmentation, using cross-entropy\,\cite{zhang2018generalized}
  and Dice\,\cite{sudre2017generalised} as loss functions. The models
  were also finetuned using the Lov\'asz loss
  function\,\cite{berman2018lovasz}.

\item \textbf{NUCTech Robot Group (NUCTech Limited Co., China) -}
  \textbf{Tasks 1 \& 2} and \textbf{1 \& 3} are resolved
  simultaneously through two unified models named ``Double
  Segmentation Network'' (DSegNet). Feature extraction of DSegNet is
  carried out by the ResNet-50\,\cite{he2016deep} backbone. These
  networks have two output channels: one outputs the document boundary
  and text segmentation zones and the other outputs document boundary
  and signature segmentation. Data augmentation has also been used in
  the training of the models with Dice\,\cite{sudre2017generalised} as
  loss function.

\item \textbf{Ocean Group (Lenovo Research and Xi'an Jiaotong
    University, China) -} \textbf{Task 1:} Ensemble framework
  consisting of two segmentation models, Mask R-CNN\,\cite{he2017mask}
  with PointRend module\,\cite{kirillov2020pointrend} and
  DeepLab\,\cite{chen2017deeplab} with decouple
  SegNet\,\cite{li2020improving}, whose outputs are merged to get the
  final result. The models were also trained with data
  augmentation. \textbf{Task 3:} A one-stage detection model is used
  to detect handwritten signature bounding box. Then a full
  convolutional encoder-decoder segmentation network combined with a
  refinement block was used to predict accurately the handwritten
  signature location.

\item \textbf{PA Group (AI Research Institute, OneConnect, China) -}
  \textbf{Tasks 1 \& 3:} The approaches are based on ResNet-34 followed
  by a ``Global Convolution Network''
  (ResNet34+GCN)\,\cite{he2016deep,zhao2017pyramid}. \textbf{Task 2:}
  The approach is based on ResNet-18 followed by a ``Pyramid Scene
  Parsing Network''
  (ResNet18+PSPNet)\,\cite{he2016deep,zhao2017pyramid}. Dice loss
  function\,\cite{sudre2017generalised} was used in the training of
  all task's models.

\item \textbf{SPDB Lab Group (Shanghai Pudong Development Bank, China)
    -} \textbf{Task 1:} The approach based on three deep networks to
  predict the mask and vote on the prediction results of each pixel:
  1) Mask R-CNN\,\cite{he2017mask} which uses
  HRNet\,\cite{sun2019deep} as backbone, HR feature pyramid networks
  (HRFPN)\,\cite{wei2020precise} and
  PointRend\,\cite{kirillov2020pointrend}; 2) Multi-scale
  DeepLabV3\,\cite{chen2017rethinking}; and 3)
  U2Net\,\cite{qin2020u2}. \textbf{Task 2:} The approach has three
  different DL models participate in the final pixel-wise voting
  prediction: 1) Multi-scale
  OCRNet\,\cite{tao2020hierarchical,yuan2019object}; 2)
  HRNet+OCRNet\,\cite{yuan2019object}; and 3)
  DeepLabV3+\,\cite{chen2018encoder}, which uses
  ResNeSt101\,\cite{zhang2020resnest} as backbone. \textbf{Task 3:}
  The approach employs object detector YOLO\,\cite{redmon2016you},
  whose cropped image output is sent to several segmentation models to
  get the final pixel-wise voting prediction: 1)
  U2Net\,\cite{qin2020u2}; 2) HRNet+OCRNet\,\cite{yuan2019object}; 3) ResNeSt+DeepLabV3 +\cite{chen2018encoder,zhang2020resnest}; and 4)
  Swin-Transformer\,\cite{liu2021swin}.

\item \textbf{SunshineinWinter Group (Xidian University, China) -}
  \textbf{Tasks 1 \& 2}: Addressed approaches use the so-called
  EfficientNet\,\cite{tan2019efficientnet} as the backbone of the
  segmentation method and the ``Feature Pyramid Networks''
  (FPN)\,\cite{lin2017feature} for object detection. Balanced
  cross-entropy\,\cite{zhang2018generalized} was used as a loss
  function. \textbf{Task 3}: The approach is based on the classical
  two-stage detection model Mask R-CNN\,\cite{he2017mask}, while the
  segmentation model is the same as the ones in Tasks 1 \& 2.

\item \textbf{USTC-IMCCLab Group (University of Science and Technology
    of China, Hefei Zhongke Brain Intelligence Technology Co. Ltd., China)
    -} \textbf{Tasks 1 \& 2}: The approached methods are based on a
  combination of a ``Cascade R-CNN''\,\cite{cai2019cascade} and
  DeepLabV3+\,\cite{chen2018encoder} for pixel prediction of document
  boundary and text zones.

\item \textbf{Wuhan Tianyu Document Algorithm Group (Wuhan Tianyu
    Information Industry Co. Ltd., China) -} \textbf{Task 1:}
  DeepLabV3\,\cite{chen2017rethinking} model was used for document
  boundary detection. \textbf{Task 2:} A two-stage approach which
  uses a DeepLabV3\,\cite{chen2017rethinking} for roughly text field
  detection and OCRNet\,\cite{liao2020real} for locating accurately
  text zones. \textbf{Task 3:} OCRNet\,\cite{liao2020real} was
  employed for handwritten signature location.
  
\end{enumerate}

\section{Results and Discussion}
\label{section:results_and_discussion}

The set of tests made available to the participants covers the three tasks of the competition. The teams were able to choose between solving one, two, or three tasks. For each Task, we had the following distribution: Task 1 with 16 teams, Task 2 with 12 teams, and Task 3 with 11 teams. Thus, 39 submissions were evaluated for each metric.

We present the results for each task of the challenge in the following tables: Table \ref{tab:challenge_1} shows the results for Task 1; in Table \ref{tab:challenge_2} are the results of Task 2; and Table \ref{tab:challenge_3} shows Task 3 results. The results of the teams were evaluated using the metrics described in Section \ref{section:competition_protocol}. We removed the SSIM metric from our analysis, since it did not prove to represent (all participants achieved results above $0.99999$). A possible reason for this behaviour is that the images resulting from the segmentation have predominantly black pixels for all Tasks. This shows that none of the teams had problems with the structural information of the image. Therefore, the metrics used to evaluate all systems were DSC and SIFT. 

For ranking the submissions, we used at first the DSC metric and as a tiebreaker criteria the SIFT one. The reason for this decision is that SIFT is invariant to rotation and scale and less sensitive to small divergences between the output images and the Ground Truth images. The results of each task will be described further and commented on the best results.

\subsection{Results of the Task 1: Document Boundary Segmentation}
In Task 1, all competitors achieved excellent results for the two metrics, DSC and SIFT. Table \ref{tab:challenge_1} shows the results and the classification of the teams. The three teams with the best classification were Ocean (1st), SPDB Lab (2nd), and PA (3rd), in this order. The first two teams achieved a result above $0.99$, and both approach a similar architecture, the Mask-RCNN model. More details on the approach of the teams can be seen in Section \ref{section:approaches}. 

\begin{table}[!htbp]
\centering
\caption{Result: 1st Challenge - Document Boundary Segmentation}
\begin{tabular}{ |p{4.5cm}|p{2.8cm}|p{2.8cm}| }
\hline
\multicolumn{3}{|c|}{Teams results and methods | Mean;$(\pm std)$} \\
\hline
Team Name & DSC$(\pm std)$ & SIFT$(\pm std)$ \\
\hline
58 CV  &  0.969594;$(\pm 0.062)$   & 0.998698;$(\pm 0.020)$\\

Ambilight  & 0.983825;$(\pm 0.034)$   & 0.999223;$(\pm 0.021)$\\

Arcanite &  0.986441;$(\pm 0.018)$   & 0.994290;$(\pm 0.037)$\\

Cinnamon AI	& 0.983389;$(\pm 0.006)$   & 0.999656;$(\pm 0.014)$\\

CUDOS &	 0.989061;$(\pm 0.002)$   & 0.999450;$(\pm 0.007)$\\

Dao Xianghu light of TianQuan & 0.987495;$(\pm 0.002)$ & 0.998966;$(\pm 0.032)$\\

DeepVision & 0.976849;$(\pm 0.008)$   & 0.999599;$(\pm 0.020)$\\

dotLAB Brazil & 0.988847;$(\pm 0.027)$   & 0.999234;$(\pm 0.009)$\\

NAVER Papago & 0.989194;$(\pm 0.002)$   & 0.999565;$(\pm 0.008)$\\

Nuctech robot &	0.986910;$(\pm 0.007)$   & 0.999891;$(\pm 0.040)$\\

Ocean & \textbf{0.990971;$(\pm 0.001)$}   & 0.999176;$(\pm 0.021)$\\

PA & \textbf{0.989830};$(\pm 0.003)$   & 0.999624;$(\pm 0.007)$\\

SPDB Lab & \textbf{0.990570};$(\pm 0.001)$   & 0.999891;$(\pm 0.003)$\\

SunshineinWinter & 0.970344;$(\pm 0.014)$   & 0.998358;$(\pm 0.017)$\\

USTC-IMCCLab & 0.985350;$(\pm 0.006)$   & 0.996469;$(\pm 0.039)$\\

Wuhan Tianyu Document Algorithm Group & 0.980272;$(\pm 0.013)$   & 0.999370;$(\pm 0.010)$\\
\hline
\end{tabular}
\label{tab:challenge_1}
\end{table}

\subsection{Results of the Task 2: Zone Text Segmentation}
Table \ref{tab:challenge_2} shows the result of Task 2. This task was slightly more complex than Task 1, as it is possible to observe slightly greater distances between the results achieved by the teams. Here the three teams with the best scores were Ambilight (1st), Dao Xianghu light of TianQuan (2nd), and USTC-IMCCLab (3rd). Here, the winner uses a model composition, HRnet+ORC+DCNv2. Based on Task 1 segmentation, the component area is cut out and then rotated to one side. Finally, the High-Resolution Net model is trained to identify the text area.

\begin{table}[!htbp]
\caption{Result: 2nd Challenge - Zone Text Segmentation}
\begin{tabular}{ |p{4.5cm}|p{2.8cm}|p{2.8cm}| }
\hline
\multicolumn{3}{|c|}{Teams results and methods | Mean;$(\pm std)$} \\
\hline
Team Name & DSC$(\pm std)$ & SIFT$(\pm std)$ \\
\hline
Ambilight & \textbf{0.926624};$(\pm  0.051)$ & 0.998829;$(\pm 0.014)$\\

Arcanite & 0.837609;$(\pm 0.069)$   & 0.997631;$(\pm 0.006)$\\

Cinnamon AI	& 0.866914;$(\pm 0.097)$   & 0.995777;$(\pm 0.009)$\\

Dao Xianghu light of TianQuan &	\textbf{0.909789};$(\pm 0.040)$ & 0.998308;$(\pm 0.004)$\\

DeepVision & 0.861304;$(\pm 0.146)$   & 0.994713;$(\pm  0.015)$\\

dotLAB Brazil & 0.887051;$(\pm 0.106)$   & 0.997471;$(\pm 0.006)$\\

Nuctech robot &	0.883612;$(\pm 0.051)$   & 0.997728;$(\pm 0.006)$\\

PA & 0.880801;$(\pm  0.063)$   & 0.997607;$(\pm 0.006)$\\

SPDB Lab & 0.881975;$(\pm  0.061)$   & 0.998220;$(\pm 0.005)$\\

SunshineinWinter &	0.390526;$(\pm 0.100)$   & 0.981881;$(\pm  0.025)$\\

USTC-IMCCLab &	\textbf{0.890434};$(\pm 0.077)$   & 0.997933;$(\pm 0.005)$\\

Wuhan Tianyu Document Algorithm Group & 0.814985;$(\pm 0.097)$   & 0.993798;$(\pm 0.011)$\\
\hline
\end{tabular}
\label{tab:challenge_2}
\end{table}

\subsection{Results of the Task 3: Signature Segmentation}
For Task 3, as reported in the Table \ref{tab:challenge_3},   results show very marked divergences. The distances between the results are much greater than those observed in the two previous tasks. The three teams that achieved the best results were SPDB Lab (1st), Cinnamon IA (2nd), and dotLAB Brasil (3rd). This task includes the lowest number of participating teams and lower rates results than tasks 1 and 2. These numbers raise the perception that segmenting handwritten signatures presents greater complexity among the 3 tasks addressed in the competition.

\begin{table}[!htbp]
\caption{Result: 3rd Challenge - Signature Segmentation}
\begin{tabular}{ |p{4.5cm}|p{2.8cm}|p{2.8cm}| }
\hline
\multicolumn{3}{|c|}{Teams results and methods | Mean;$(\pm std)$} \\
\hline
Team Name & DSC$(\pm std)$ & SIFT$(\pm std)$ \\
\hline
Ambilight  & 0.812197;$(\pm 0.240)$   & 0.839524;$(\pm 0.360)$\\

Arcanite & 0.627745;$(\pm 0.291)$   & 0.835515;$(\pm 0.359)$\\

Cinnamon AI	& \textbf{0.841275};$(\pm 0.226)$   & 0.848724;$(\pm 0.353)$\\

Dao Xianghu light of TianQuan & 0.795448;$(\pm 0.262)$   & 0.839367;$(\pm 0.361)$\\

dotLAB Brazil & \textbf{0.837492};$(\pm 0.412)$   & 0.852816;$(\pm 0.324)$\\

Nuctech robot & 0.538511;$(\pm 0.316)$   & 0.842416;$(\pm 0.351)$\\

Ocean & 0.768071;$(\pm 0.286)$   & 0.838772;$(\pm 0.362)$\\

PA & 0.407228;$(\pm 0.314)$ & 0.836094;$(\pm 0.356)$\\

SPDB Lab &	\textbf{0.863063};$(\pm 0.245)$   & 0.839501;$(\pm 0.361)$\\

SunshineinWinter &	0.465849;$(\pm 0.283)$   & 0.835916;$(\pm 0.357)$\\

Wuhan Tianyu Document Algorithm Group & 0.061841;$(\pm 0.765)$   & 0.589850;$(\pm 0.458)$\\
\hline
\end{tabular}
\label{tab:challenge_3}
\end{table}

The winning group, SPDB Lab, used a strategy that involves several deep models.
The Yolo model was used by this group to detect the bounding box (bbox) of the handwritten signature area. Then, the bbox is filled with 5 pixels so that the overlapping bounding box is later merged.
These images are sent to various segmentation models for prediction, which are U2Net, HRNet+OCRNet, ResNeSt+Deeplabv3plus, Swin-Transformer.
The results of the segmentation of various models are merged by the weighted sum, and then the results are pasted back to the original position according to the recorded location information.

\subsection{Discussion}

All tasks proposed in this competition received  interesting strategies and some of them achieved promising results for the problems posed by the taks. 

Task 1 was completed with excellent results, all above $0.96$ for DSC metric. With this result of the competition, it is possible to affirm that Document Boundary Segmentation is no longer a challenge for computer vision due to the power of deep learning architectures. 

The results of Task 2 show that the challenge for segmenting text zones needs an additional effort of the research community to get a definitive solution, since a minor error in this task may produce substantial errors when recognizing the texts in those documents.

Task 3, Handwritten signature segmentation, showed a more complex task than the other two presented in this competition. Despite the good results presented specially by Top 3 systems, they are far from the rates of page detection algorithms. Therefore, if we need the signature pixels to further verify its similarity against a reference signature, there is still room for new strategies and improvements in this task.

All groups used deep learning models, most of which made use of encoder-decoder architectures like the U-net. Another characteristic worth noting is the strategy of composing several deep models working together or sequentially to detect and segment regions of interest.

In the following, we can see some selected results of the 3 winners teams of each task.

\begin{figure}[!htb]
  \centering
  \includegraphics[width=.7\textwidth]{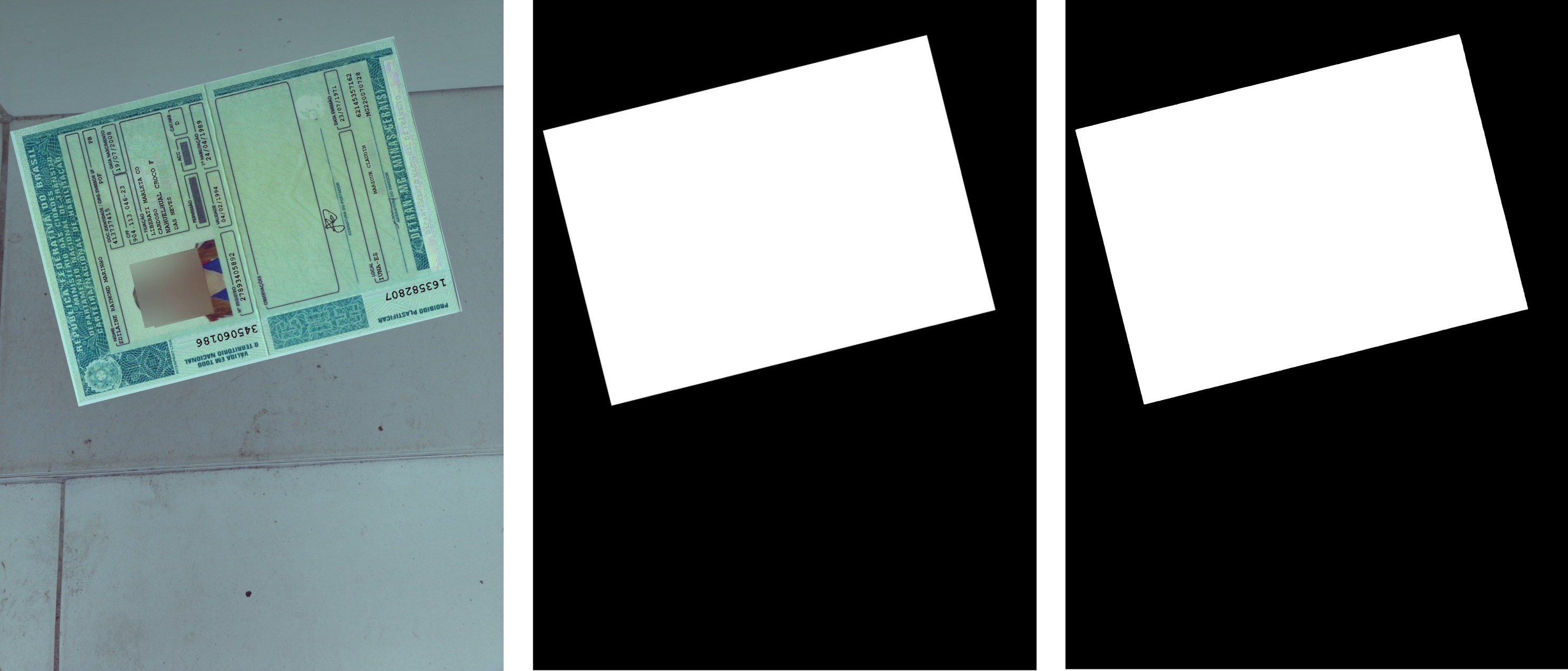}
  \caption{\textbf{Sample of the Ocean team's result - Task 1}. On the left the input image, in the centre the ground-truth image and on the right the resulting output image.}
  \label{fig:task1_ocean}
\end{figure}

\begin{figure}[!htb]
  \centering
  \includegraphics[width=.7\textwidth]{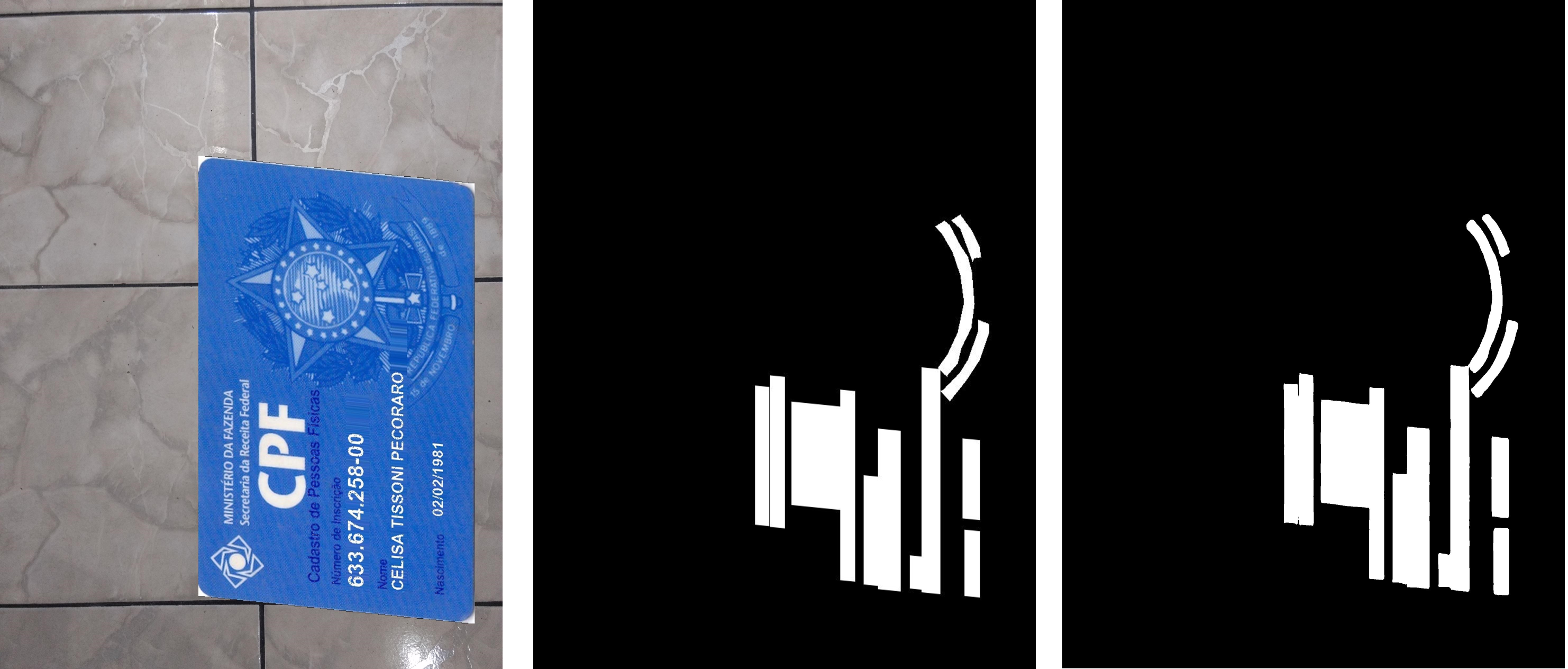}
  \caption{\textbf{Sample of the Ambilight team's result - Task 2}. On the left the input image, in the centre the ground-truth image and on the right the resulting output image.}
  \label{fig:task2_ambilight}
\end{figure}

\begin{figure}[!htb]
  \centering
  \includegraphics[width=.7\textwidth]{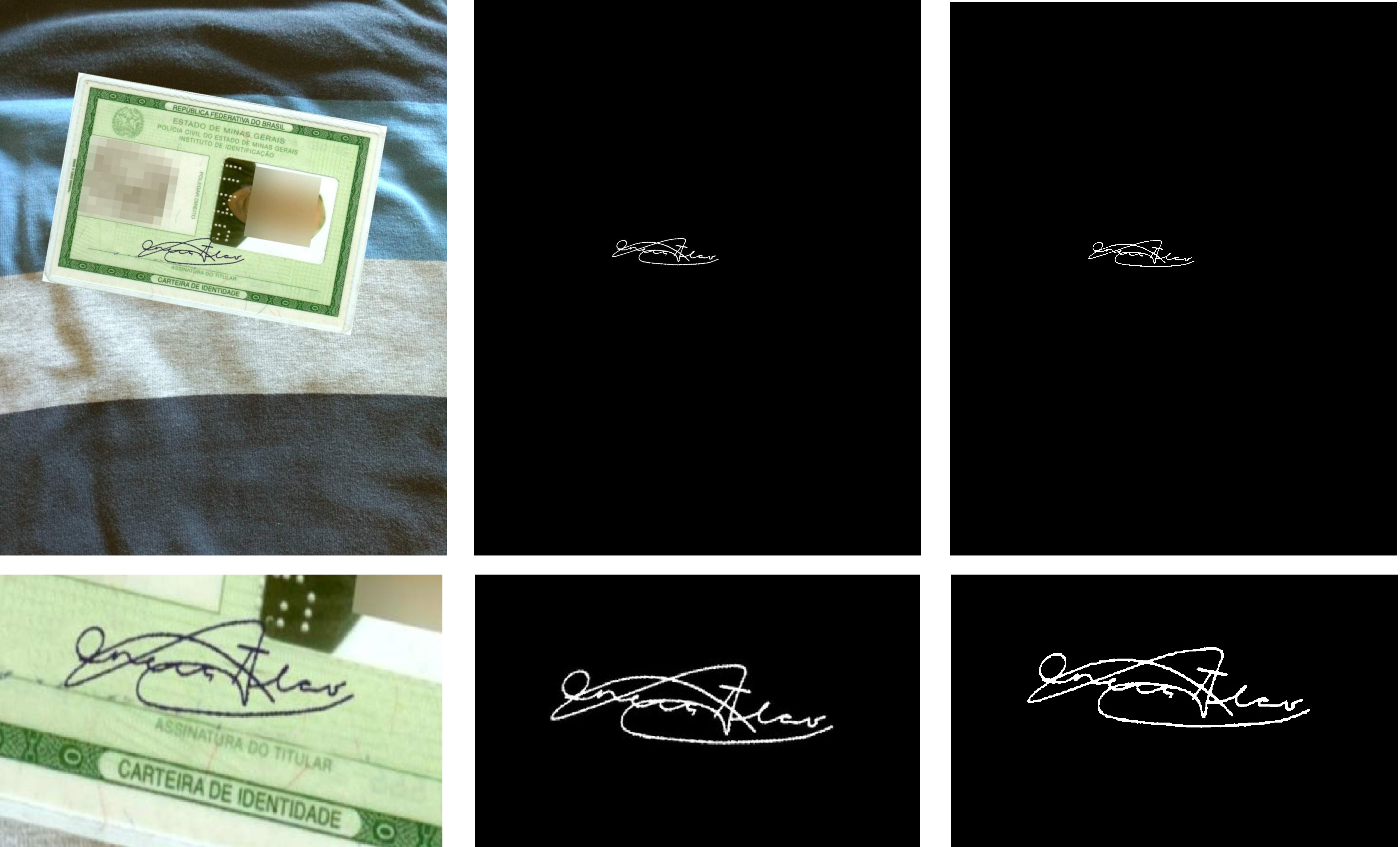}
  \caption{\textbf{Sample of the SPDB Lab team's result - Task 3}. On the left the input image, in the centre the ground-truth image and on the right the resulting output image. Just below the original images are the regions of enlarged signatures for better viewing.}
  \label{fig:task3_SPDBLab}
\end{figure}

\section{Conclusion}

The models proposed by the competitors were based on consolidated state-of-the-art architectures. This increased the level and complexity of the competition. Strategies such as using one task as the basis for another task were also successfully applied by some groups. 

Nevertheless, we are worried about the response time of such approaches which is important in the case of real-time applications. In addition, one must also consider the application for embedded systems and mobile devices with limited computational resources. A possible future competition may include in the scope these aspects. 
 
 To conclude, we see that there is still room for other strategies and solutions for the tasks of text and signature segmentation proposed in this competition. We expect the datasets produced in this contest (\url{https://icdar2021.poli.br/}) can be used by the research community to improve their proposals for such tasks.

\section*{Acknowledgment}
This study was financed in part by: Coordenação de Aperfeiçoamento de Pessoal de Nível Superior - Brasil (CAPES) - Finance Code 001, and CNPq - Brazilian research agencies.

\bibliographystyle{splncs04}
\bibliography{main}

\end{document}